\def\year{2021}\relax
\documentclass[letterpaper]{article} 
\usepackage{aaai21}  
\usepackage{times}  
\usepackage{helvet} 
\usepackage{courier}  
\usepackage[hyphens]{url}  
\usepackage{graphicx} 
\usepackage{natbib}  
\usepackage{caption} 
\usepackage{amsmath}
\usepackage{array}
\usepackage{mdwmath}
\usepackage{mdwtab}
\usepackage{eqparbox}
\usepackage{stackrel}
\usepackage{bm}
\usepackage{multirow}
\usepackage{threeparttable}
\usepackage{booktabs}
\usepackage{amssymb}
\usepackage{diagbox}
\usepackage{subcaption}
\usepackage{algorithm}
\usepackage{algorithmic}
\usepackage{amsthm}
\usepackage{boldline}
\usepackage{cleveref}
\usepackage[switch]{lineno}

\title{OPQ: Compressing Deep Neural Networks with One-shot Pruning-Quantization}
\author{
    Peng Hu\textsuperscript{\rm 1, 2}, Xi Peng\textsuperscript{\rm 2}, Hongyuan Zhu\textsuperscript{\rm 1}, Mohamed M. Sabry Aly\textsuperscript{\rm 3}, Jie Lin\textsuperscript{\rm 1,}\thanks{Corresponding author: Jie Lin.}\\
}
\affiliations{
    \textsuperscript{\rm 1}Institute for Infocomm Research, Agency for Science, Technology and Research, Singapore\\
    \textsuperscript{\rm 2}College of Computer Science, Sichuan University, Chengdu 610065, China\\
    \textsuperscript{\rm 3}Nanyang Technological University, Singapore\\
    \{penghu.ml, pengx.gm, hongyuanzhu.cn\}@gmail.com; msabry@ntu.edu.sg; lin-j@i2r.a-star.edu.sg
}

\begin{document}
    \maketitle
	\begin{abstract}
		As Deep Neural Networks~(DNNs) usually are overparameterized and have millions of weight parameters, it is challenging to deploy these large DNN models on resource-constrained hardware platforms, \textit{e.g.}, smartphones. Numerous network compression methods such as pruning and quantization are proposed to reduce the model size significantly, of which the key is to find suitable compression allocation (\textit{e.g.}, pruning sparsity and quantization codebook) of each layer. 
		Existing solutions obtain the compression allocation in an iterative/manual fashion while finetuning the compressed model, thus suffering from the efficiency issue. Different from the prior art, we propose a novel One-shot Pruning-Quantization (OPQ) in this paper, which analytically solves the compression allocation with pre-trained weight parameters only. During finetuning, the compression module is fixed and only weight parameters are updated. To our knowledge, OPQ is the first work that reveals pre-trained model is sufficient for solving pruning and quantization simultaneously, without any complex iterative/manual optimization at the finetuning stage.
		Furthermore, we propose a unified channel-wise quantization method that enforces all channels of each layer to share a common codebook, which leads to low bit-rate allocation without introducing extra overhead brought by traditional channel-wise quantization. Comprehensive experiments on ImageNet with AlexNet/MobileNet-V1/ResNet-50 show that our method improves accuracy and training efficiency while obtains significantly higher compression rates compared to the state-of-the-art.
	\end{abstract}
	
	\section{Introduction}
	In recent years, Deep Neural Networks (DNNs) have achieved great success in various applications, such as image classification~\cite{he2016deep,li2020neural} and object detection~\cite{he2017mask}. However, the massive computation and memory cost hinder their deployment on lots of popular resource-constraint devices, \textit{e.g.}, mobile platform~\cite{han2015deep,li2019additive}. 
	The well-known over-parameterization of DNNs shows the theoretical basis to build lightweight versions of these large models~\cite{zhang2020one,yang2020cars,li2020neural}. Therefore, DNN model compression without accuracy loss has gradually become a hot research topic. 
	
	
	
	Network pruning is one of the popular compression techniques, by removing redundant weight parameters to slim the DNN models without losing any performance, \textit{e.g.}, fine-grained weight pruning~\cite{guo2016dynamic}, filter pruning~\cite{luo2017thinet}, etc. One key problem for weight pruning is to determine which neuron is deletable. Recently, the lottery ticket hypothesis~\cite{frankle2018lottery} proves that the removable neurons can be directly determined by the pre-trained model only, without updating the pruning masks during finetuning~(\textit{a.k.a.}, one-shot weight pruning). 
	Obviously, it is promising to explore whether this one-shot mechanism generalizes to other compression techniques, \textit{e.g.}, one-shot quantization. 
	Like one-shot pruning, one-shot quantization is expected to support finetuning the quantized model without updating the quantizers derived from the pre-trained model. 
	
\begin{figure}[t]
		\begin{subfigure}[b]{0.475\linewidth}
			\includegraphics[height=0.75\linewidth]{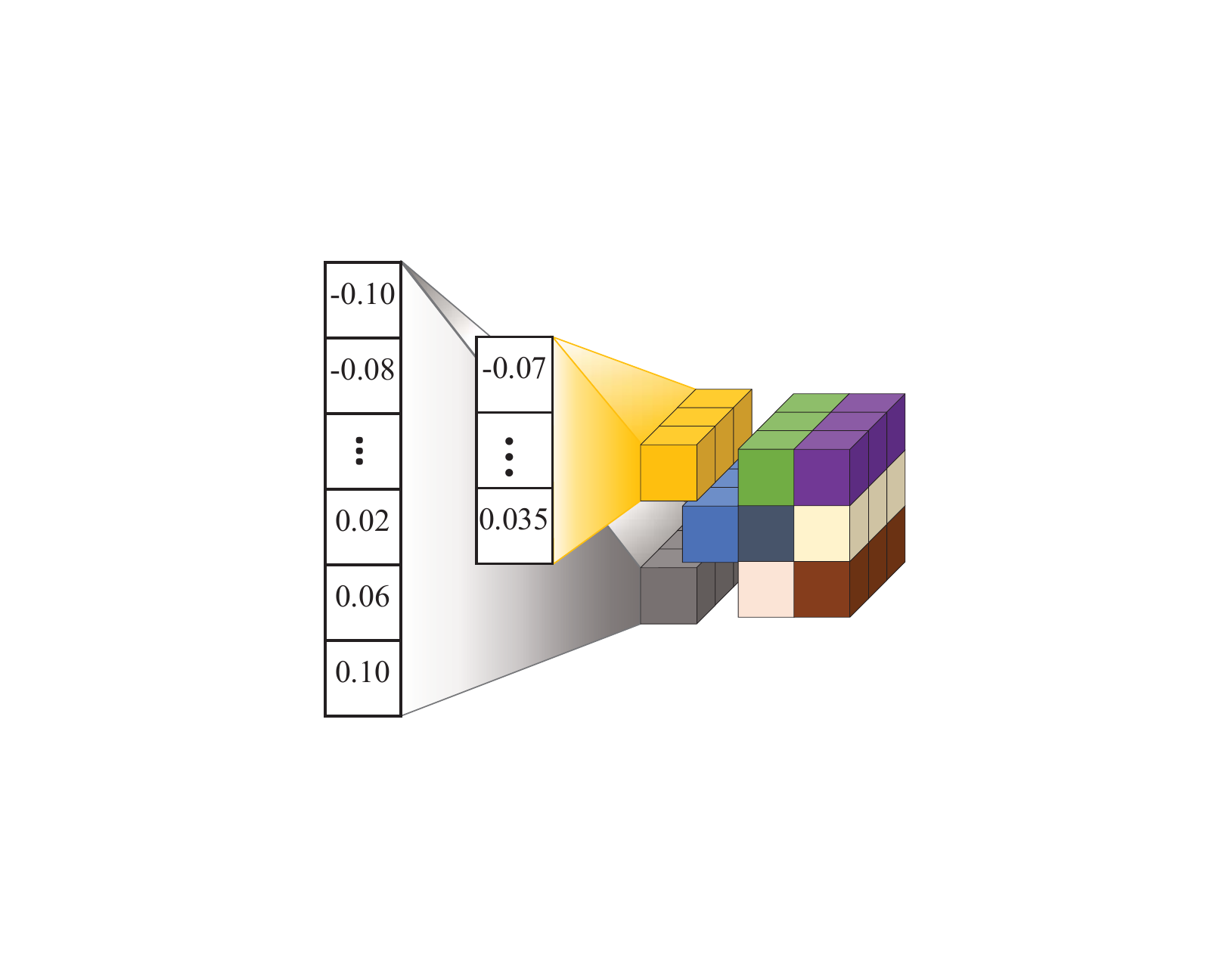}
			\captionsetup{font={small}}
			\caption{}\label{fig:channel_wise_quantization}
		\end{subfigure}
		\hspace{2mm}
		\begin{subfigure}[b]{0.475\linewidth}
			\centering
			\includegraphics[height=0.75\linewidth]{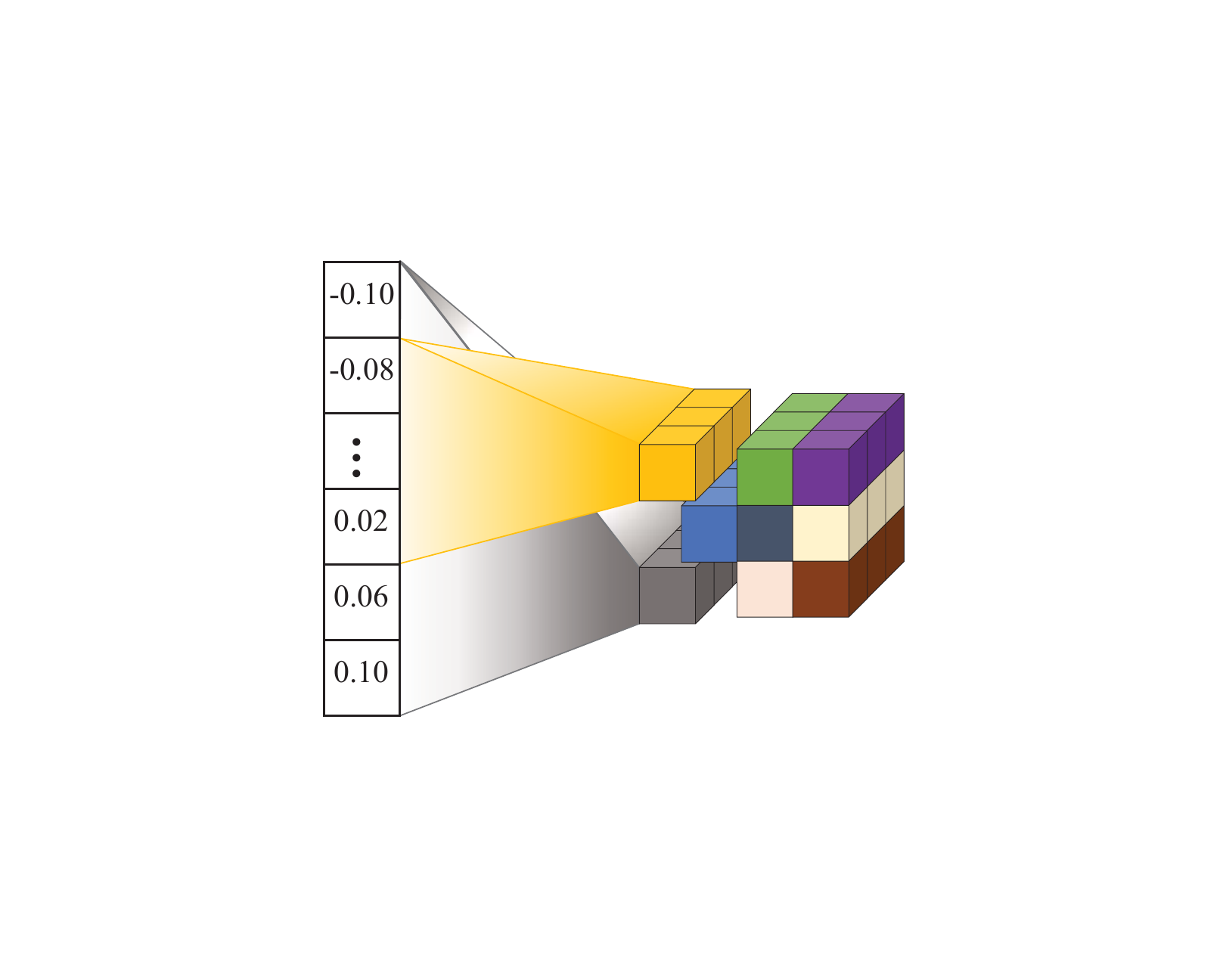}
			\captionsetup{font={small}}
			\caption{}\label{fig:proposed_channel_wise_quantization}
		\end{subfigure}
		\caption{Difference between (\subref{fig:channel_wise_quantization}) existing channel-wise quantization and (\subref{fig:proposed_channel_wise_quantization}) the proposed unified quantization. Cubes with the same color denote a channel. In (\subref{fig:channel_wise_quantization}), each channel of a layer has its individual quantization codebook. On the contrary, in (\subref{fig:proposed_channel_wise_quantization}) all channels of a layer share a common codebook.}
		\label{fig:difference}
	\end{figure}

	Network quantization opts to represent weight parameters using fewer bits instead of 32-bit full precision~(FP32), leading to reduced memory footprint and computation cost while maintaining accuracy as much as possible.
	Quantization could be roughly classified as layer-wise~\cite{he2018learning,wang2019haq} and channel-wise~\cite{krishnamoorthi2018quantizing} methods. For layer-wise quantization, the weights of each layer are quantized by a common layer-specific quantizer. 
	Conversely, for channel-wise quantization, each channel of a layer has its own channel-specific quantizer, \textit{i.e.}, multiple different channel-specific quantizers are required for a layer as illustrated in \Cref{fig:channel_wise_quantization}. With comparable accuracy, the fine-grained channel-wise quantization usually achieves a higher compression rate than the coarse-grained layer-wise quantization, while at the expense of extra overhead introduced (\textit{i.e.}, channel-wise codebooks, etc.) which in turn increases the difficulty for hardware implementation~\cite{nagel2019data,cai2020zeroq}.
	
	To further maximize the compression rate, it is desirable to simultaneously conduct pruning and quantization on the DNN models~\cite{wang2020apq,yang2020automatic}. However, the pruning-quantization strategy is more complex than either pruning or quantization alone. 
	In other words, it is almost impossible for manually tuning the pruning ratios and quantization codebooks at fine-grained levels. To address this issue, recent methods adopt certain optimization techniques, \textit{e.g.}, Bayesian optimization~\cite{tung2020deep}, Alternating Direction Method of Multipliers (ADMM)~\cite{yang2020automatic} to iteratively update the compression allocation~(\textit{i.e.}, pruning ratios and quantization codebooks) and weights of the compressed models. Although these pruning-quantization methods achieve promising results, optimizing the compression allocation parameters introduces considerable computational complexity at the finetuning stage.

	\begin{figure*}[!thb]
		\centering
		\includegraphics*[width=0.95\textwidth]{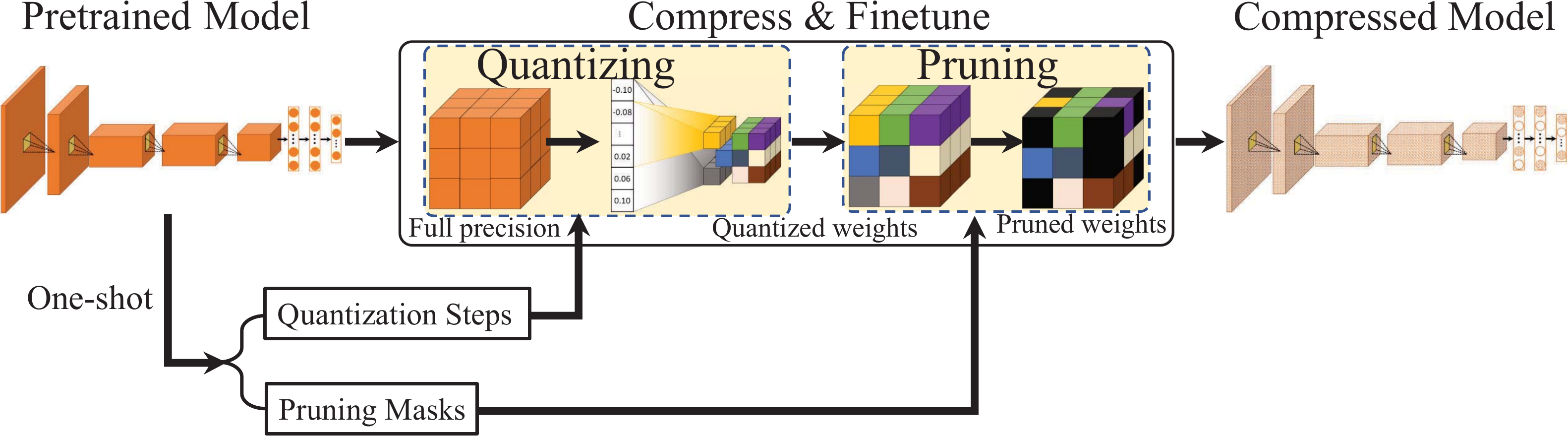}
		\caption{The pipeline of our method. Channels of each layer share the same quantizer~(\textit{i.e.}, the same codebook). Cubes in black color indicate pruned weights. Given a pre-trained model, the pruning masks $\{ \mathbf{M}_{i} \}_{i=1}^{L}$ (see \Cref{pruning}) and quantization steps $\{ \Delta_{i} \}_{i}^{L}$ (see \Cref{quantization}) are analytically derived in one-shot and fixed while finetuning the compressed model.}
		\label{fig:framework}
	\end{figure*}
	
	To address the aforementioned problems, we proposed a novel One-shot Pruning-Quantization method~(OPQ) to compress DNN models. 
	Given a pre-trained model, a unified pruning error is formulated to calculate layer-wise pruning ratios analytically which subsequently derive layer-wise magnitude-based pruning masks.
	Then we compute a common channel-wise quantizer for each layer analytically by minimizing the unified quantization error of the pruned model.
	At the finetuning stage, the pruning mask and the channel-wise quantizer of each layer are fixed and only the weight parameters are updated in order to compensate for accuracy loss incurred by the model compression error.
	Thus, our method provides an analytical solution to obtain suitable pruning-quantization allocations in a one-shot manner, avoiding iterative, variable and complex optimization of the compression module during finetuning.
	Moreover, the proposed unified channel-wise quantization enforces all channels of each layer to share a common quantization codebook as illustrated in \Cref{fig:difference}. Unlike traditional channel-wise quantization methods, our unified quantization does not introduce any overheads (\textit{i.e.}, channel-wise scales and offsets).
	Thus, our method embraces the benefits of both the fine-grained channel-wise quantization and the coarse-grained layer-wise quantization simultaneously.
	
	The main contributions of this work are three-fold:
	\begin{enumerate}
		\item A One-shot Pruning-Quantization method for compression of DNNs. To our knowledge, this is the first work reveals that pre-trained model is sufficient for solving pruning and quantization simultaneously, without any iterative and complex optimization during finetuning.
		\item A Unified Channel-wise Quantization method remarkably reduces the number of channel-wise quantizers for each layer, and thus avoids the overhead brought by the traditional channel-wise quantization.
		\item Extensive experiments prove the effectiveness of the one-shot pruning-quantization strategy. On AlexNet, MobileNet-V1, and ResNet-50, our method boosts the state-of-the-art in terms of accuracy and compression rate by a large margin.
	\end{enumerate}
	
	\section{Related Works}
	In this section, we briefly review the most related works from the following three aspects: pruning, quantization, and pruning-quantization methods.
	
	\subsection{Pruning Methods}
	Pruning aims at removing unimportant weights/neurons to slim the over-parameterized DNN models.
	Since fully-connected layers contain the dominant number of parameters in early stage of the classical DNN models, research works focus on removing the neurons in the fully-connected layers~\cite{yang2015deep,cheng2015exploration}. Follow-up works attempt to sparsify the weights in both convolutional and fully-connected layers~\cite{srinivas2015data,guo2016dynamic,zhu2017prune}, \textit{e.g.}, with tensor low rank constraints, group sparsity~\cite{zhou2016less}, constrained Bayesian optimization~\cite{chen2018constraint}, etc.
	Pruning methods can be roughly grouped into unstructured pruning (\textit{e.g.}, weights)~\cite{guo2016dynamic} or structured pruning (\textit{e.g.}, filters)~\cite{luo2017thinet,he2020learning}. 
    The former shrinks network size by simply defining hard threshold for pruning criteria such as magnitude of weights. The latter removes the sub-tensors along a given dimension in the weight tensors with pruning criteria such as L1/L2 norm of filters. Structured pruning is compatible with existing hardware platforms, while unstructured pruning requires the support of hardware architecture with specific design. On the other side, unstructured pruning usually achieves much higher compression rate than structured pruning at comparable accuracy, which is an attractive feature for embedded systems that with extreme-low resource constraints.

	\subsection{Quantization Methods}
	Quantization enforces the DNN models to be represented by low-precision numbers instead of 32-bit full precision representation, leading to smaller memory footprint as well as lower computational cost. In recent years, numerous quantization methods are proposed to quantize the DNN models to as few bits as possible without accuracy drop, \textit{e.g.}, uniform quantization with re-estimated statistics in batch normalization and estimation error minimization of each layer's response~\cite{he2018learning}, layer-wise quantization with reinforcement learning~\cite{wang2019haq}, channel-wise quantization~\cite{banner2019post}, etc. The extreme case of quantization is binary neural networks, which represent the DNN weights with binary values. For instance, Rastegari \textit{et al.} adopted binary values to approximate the filters of DNN, called Binary-Weight-Networks~\cite{wu2016quantized}.

	
	\subsection{Pruning-quantization Methods}
	Obviously, both pruning and quantization can be simultaneously conducted to boost the compression rate. Han \textit{et al.} proposed a three-stage compression pipeline (\textit{i.e.}, pruning, trained quantization and Huffman coding) to reduce the storage requirement of DNN models~\cite{han2015deep}. In-parallel pruning-quantization methods are proposed to compress DNN models to smaller size without losing accuracy. Specifically, Tung \textit{et al.} utilized Bayesian optimization to prune and quantize DNN models in parallel during fine-tuning~\cite{tung2020deep}. In \cite{yang2020automatic}, Yang \textit{et al.} proposed an optimization framework based on Alternating Direction Method of Multipliers (ADMM) to jointly prune and quantize the DNNs automatically to meet target model size.
	For the aforementioned methods, the pruning masks and quantizers for all layers are manually set or iteratively optimized at finetuning stage. How to directly obtain the desirable compression module from the pre-trained models without iterative optimization at finetuning stage is less explored in earlier works.

	\section{The Proposed Method}
	\subsection{Preliminaries}
	We first provide notations and definitions for neural network compression, \textit{i.e.}, pruning and quantization. Let $\mathcal{W} = \{ \mathbf{W}_{i} \}_{i = 1}^{L}$ be the FP32 weight tensors of a $L$-layer pre-trained model, where $\mathbf{W}_{i}$ is the weight parameter tensor of the $i$-th layer. To simplify the presentation, let $W_{ij}~(j = 1, 2, \cdots, N_{i})$ be the $j$-th element of $\mathbf{W}_{i}$ and $W_{ijk}~(k = 1, 2, \cdots, N_{ij})$ be $k$-th element from the $j$-th channel of the $i$-th layer, where $N_{i}$ is the number of weight parameters in $\mathbf{W}_{i}$ and $N_{ij}$ is the number of weight parameters from the $j$-th channel of $i$-th layer. The weights of each layer satisfy a symmetrical distribution around zero such as Laplace distribution~\cite{ritter2018scalable,banner2019post}. We define a probability density function $f_{i}(x)$ for the $i$-th layer. Different from the prior art~\cite{han2015deep,yang2020automatic,tung2020deep}, given a target pruning rate and quantization bit-rate, our method analytically solves the pruning-quantization allocations (\textit{i.e.}, pruning masks $\{ \mathbf{M}_{i} \}_{i=1}^{L}$ and quantization steps $\{ \Delta_{i} \}_{i}^{L}$) with pre-trained weight parameters only in one-shot, meaning that the pruning masks and quantization steps are fixed while finetuning the weights of the compressed model. Given a pre-trained model, the pruning-quantization is conducted on its weight parameters $\mathbf{W}$ by $\hat{\mathbf{W}} = \mathbf{M} \circ \left(\Delta \lfloor \frac{\mathbf{W}}{\Delta} \rceil \right)$, where $\circ$ is the Hadamard product, $\lfloor \cdot \rceil$ is the round operator, and $\Delta \lfloor \frac{\mathbf{W}}{\Delta} \rceil$ represents the quantized weights with quantization step $\Delta$. For finetuning the compressed model, the Straight-Through Estimator~(STE)~\cite{bengio2013estimating} is applied on $\Delta \lfloor \frac{\mathbf{W}}{\Delta} \rceil$ to compute the gradients in the backward pass. 
	
	The pipeline of the proposed method is shown in \Cref{fig:framework} and \Cref{alg:method}. The details on solving $\{ \mathbf{M}_{i} \}_{i=1}^{L}$ and $\{ \Delta_{i} \}_{i}^{L}$ will be given in the following sections.
	
	\begin{algorithm}[htb]
		\caption{Optimization process of our method}
		\label{alg:method}
		\textbf{Input:} A pre-trained FP32 model with $L$ layers, objective pruning rate $p^{*}$, objective quantization bitwidth $B$, batch size $N_{b}$, and maximum epoch number $N_{e}$. \\
		\textbf{Output:} Finetuned compressed model.
		
		\begin{algorithmic}[1]
			\STATE Compute the pruning masks $\{ \mathbf{M}_{i} \}_{i=1}^{L}$ for all layers (see \Cref{pruning}).
			\STATE Calculate the qunatization steps $\{ \Delta_{i} \}_{i}^{L}$ for all layer (see \Cref{quantization}).
			\FOR{$1, 2, \cdots, N_{e}$}
			\REPEAT
			\STATE Randomly sample a minbatch from the training set.
			\STATE Compress the weights using $\{ \Delta_{i} \}_{i}^{L}$ and $\{ \mathbf{M}_{i} \}_{i=1}^{L}$ for the model.
			\STATE Forward propagate with the pruned and quantized weights, and compute the cross entropy loss.
			\STATE Update the model weights with descending their stochastic gradient.
			\UNTIL{all samples selected}
			\ENDFOR
		\end{algorithmic}
	\end{algorithm}
	
	\subsection{Unified Layer-wise Weight Pruning}\label{pruning}
	In this section, we present a general unified formulation to prune the weights $\mathcal{W}$ of all layers given a pre-trained neural network model. The pruning problem aims to find which weights could be removed. To simplify the formulation, we reformulate the problem as finding the pruning ratios of all layers $\{ p_{i} \}_{i=1}^{L}$, where $p_{i}$ is the percentage of weights with small magnitude (\textit{i.e.}, minor absolute values around zero) to be removed in the $i$-th layer. Specifically, it removes the weights in a symmetric range $[-\beta_{i}, \beta_{i} ]$ around zero for the $i$-th layer like \cite{tung2020deep}, where $\beta_{i}$ is a positive scalar value. Thus, the pruning rate of the whole model can be calculated by
	\begin{equation}\label{eq:p}
	\resizebox{.43\textwidth}{!}{$
	\displaystyle p = \frac{1}{N} \sum\limits_{i=1}^{L} \int_{-\beta_{i}}^{\beta_{i}} N_{i} f_{i}(x)dx = \frac{2}{N} \sum\limits_{i=1}^{L} \int_{0}^{\beta_{i}} N_{i} f_{i}(x)dx.$}
	\end{equation}
	Accordingly, the pruning error of the $i$-th layer can be formulated as follows:
	\begin{equation}\label{eq:obj_i}
	\mathcal{L}_{i}^{\beta} = \sum_{j=1}^{N_{i}} \left( W_{ij} \right)^{2} \bigg\rvert_{|W_{ij}| \leqslant \beta_{i}} = 2 \int_{0}^{\beta_{i}} N_{i} x^{2} f_{i}(x)dx.
	\end{equation}
	
	Obviously, we aim to minimize the errors caused by weight pruning, with the objective that the pruned model is as consistent as possible with the original model. To achieve the goal, we have the following pruning objective function:
	\begin{equation}\label{eq:obj}
	\begin{split}
	\beta^{*}_{1}, \cdots, \beta^{*}_{L} &= \stackrel[\beta_{1}, \beta_{2}, \cdots, \beta_{L}]{}{\mathop{\arg\min}} \frac{1}{N} \sum_{i = 1}^{L} \mathcal{L}_{i}^{\beta} \\
	&= \stackrel[\beta_{1}, \beta_{2}, \cdots, \beta_{L}]{}{\mathop{\arg\min}} \frac{2}{N} \sum_{i=1}^{L} \int_{0}^{\beta_{i}} N_{i} x^{2} f_{i}(x)dx.
	\end{split}
	\end{equation}
	For a given objective pruning rate $p^{*}$, we could solve the minimizing problem of \Cref{eq:obj} through the Lagrange multiplier. With the Lagrangian with a multiplier $\lambda$, we can obtain:
	\begin{equation}
	\resizebox{.43\textwidth}{!}{ %
		$\begin{split}
		\mathcal{L}(\beta_{1}, \cdots, \beta_{L}, \lambda) &= \sum_{i=1}^{L} \frac{2}{N} \int_{0}^{\beta_{i}} N_{i} x^{2} f_{i}(x)dx \\
		&\quad- \lambda \left( \frac{2}{N} \sum_{i=1}^{L} \int_{0}^{\beta_{i}} N_{i} f_{i}(x)dx - p^{*}\right).
		\end{split}$}
	\end{equation}
	In order to solve for $\beta_{i}$, we set the partial derivative of the Lagrangian function $\mathcal{L}(\beta_{1}, \cdots, \beta_{L}, \lambda)$ with respect to $\beta_{i}$ as zero.
	Then, we can obtain $\beta_{1}^{*} = \beta_{2}^{*} = \cdots = \beta_{L}^{*} = \sqrt{\lambda}$. Like the solution process of $\{\beta_{i}\}_{i = 1}^{L}$, we can obtain the $\lambda$ through setting the partial derivative of the Lagrangian function $\mathcal{L}(\beta_{1}, \cdots, \beta_{L}, \lambda)$ with respect to $\lambda$ as zero:
	\begin{equation}\label{lambda1}
	\begin{split}
	\frac{\partial \mathcal{L}(\beta_{1}, \cdots, \beta_{L}, \lambda)}{\partial \lambda} 
	&= \frac{2}{N} \sum_{i=1}^{L} \int_{0}^{\sqrt{\lambda}} N_{i} f_{i}(x)dx - p^{*} \\
	&= 0.
	\end{split}
	\end{equation}
	Since $f_{i}(x)$ could be a Laplace probability density function $\frac{1}{2\tau_{i}} e^{-\frac{|x|}{\tau_{i}}}$~\cite{banner2019post}, the probability distribution function could be obtained by $F(x) = \int_{0}^{x} f_{i}(y) dy =  \int_{0}^{x} \frac{1}{2\tau_{i}} e^{-\frac{|y|}{\tau_{i}}} dy = -\frac{1}{2} e^{-\frac{y}{\tau_{i}}}\Big\rvert_{0}^{x} $. Then, \Cref{lambda1} can be reformulated as follows:
	\begin{equation}\label{lambda2}
	\begin{split}
	\frac{\partial \mathcal{L}(\beta_{1}, \cdots, \beta_{L}, \lambda)}{\partial \lambda} 
	&\approx \frac{1}{N} \sum_{i = 1}^{L} N_{i} \left(1 - e^{-\frac{\sqrt{\lambda}}{\tau_{i}}}\right) - p^{*} \\
	&= 0.
	\end{split}
	\end{equation}
	The well-known Levenberg-Marquardt algorithm could be used to fit $F(x)$ for the weights at each layer in order to derive the layer-wise scalar parameter $\tau_{i}$.
	Moreover, we use Newton-Raphson method to solve \Cref{lambda2} in order to obtain $\lambda$, then $\{ \beta_{i} \}_{i = 1}^{L}$ could be derived with the obtained $\lambda$. $\{ \beta_{i} \}_{i = 1}^{L}$ can be used to calculate the pruning ratio of each layer by $p_{i}=\int_{-\beta_{i}}^{\beta_{i}} N_{i} f_{i}(x)dx$, and meanwhile derive the binary pruning mask $\mathbf{M}_{i} \in \{ 0, 1\}^{|\mathbf{W}_{i}|}$ via magnitude-based thresholding. 
	The pruning mask $\mathbf{M}_{i}$ is used to remove unimportant weights from the $i$-th layer by $\mathbf{M}_{i} \circ \mathbf{W}_{i}$, where $\circ$ is the Hadamard product. 
	The remaining unpruned weights are sparse tensors that could be stored by leveraging index differences and encoded with lower bits, one can refer to~\cite{han2015deep} for more implementation details.
	
	
	\subsection{Unified Channel-wise Weight Quantization}\label{quantization}
	Different from existing per-channel weight quantization \cite{krishnamoorthi2018quantizing,banner2019post}, we enforce all channels of a layer to share a common scale factor and offset~(\textit{i.e.}, a common codebook), so that it does not introduce additional overheads (\textit{i.e.}, channel-wise codebooks). To this end, we perform uniform quantization on the unpruned weights of all channels with a unified quantization step, \textit{i.e.}, a unified codebook. That is to say, all channels of a layer share a common quantization step $\Delta$ between two adjacent quantized bins/centers. 
	Let $K_{ij}$ denote the number of bins assigned to the $j$-th channel of the $i$-th layer, which is the number of quantization centers required to store the unpruned weights of the channel. Denoting a positive real value $\alpha_{ij}$ as the maximum for the $j$-th channel (\textit{i.e.}, $W_{ijk} \in [-\alpha_{ij}, \alpha_{ij}]$), the range $[-\alpha_{ij}, \alpha_{ij}]$ can be partitioned to $K_{ij}$ equal quantization bins. Therefore, the number of quantization bins $K_{ij}$ is established as follows:
	\begin{equation}
	K_{ij} = \left\lfloor \frac{2\alpha_{ij}}{\Delta_{i}} \right\rceil,
	\end{equation}
	where $\lfloor \cdot \rceil$ is the round operator which rounds a value to the nearest integer, $\alpha_{ij} = \max\left\{|W_{ijk}|~\big\rvert k = 1, 2, \cdots, \bar{N}_{ij} \right\}$, $|\cdot|$ is the absolute value operator, $W_{ijk}$ is the $k$-th weight from the $j$-th channel of the $i$-th layer, $\bar{N}_{ij}$ is the number of unpruned weights in the $j$-th channel of the $i$-th layer. Obviously, the average number of quantized bins required for the $i$-th layer is computed as follows:
	\begin{equation}
	K_{i} = \frac{1}{\bar{N}_{i}} \sum_{j = 1}^{C_{i}} \bar{N}_{ij} K_{ij},
	\end{equation}
	where $\bar{N}_{i} = \sum_{j = 1}^{\bar{N}_{i}} \bar{N}_{ij}$ is the number of all unpruned weights in the $i$-th layer. Thus, the average number of bins for the whole model could be formulated as follows:
	\begin{equation}\label{bits}
	\resizebox{.43\textwidth}{!}{ $%
	\displaystyle \frac{1}{\bar{N}} \sum\limits_{i=1}^{L} K_{i} N_{i} \int_{\beta_{i}}^{+\infty} f_{i}(x) dx 
	= \frac{1}{\bar{N}} \sum\limits_{i=1}^{L} \sum\limits_{j = 1}^{C_{i}} \bar{N}_{ij} K_{ij}  = 2^{B},
	$}
	\end{equation}
	where $B$ is the objective number of bits required to store the unpruned weights for all layers, and $\bar{N} = \sum_{i = 1}^{L} \bar{N}_{i}$ is the number of all unpruned weights in the model. With the quantization step $\Delta_{i}$, each weight of the corresponding layer could be uniformly quantized as certain discrete levels. The channels from the same layer share a common quantization function denoted as:
	\begin{equation}\label{Q}
	Q_{i}(x) = \text{sgn}(x) \Delta_{i} \bigg\lfloor \frac{|x|}{\Delta_{i}} \bigg\rceil,
	\end{equation}
	where $\text{sgn}(x)$ is the sign function. Then, the mean-square-errors caused by quantization can be formulated as:
	\begin{equation}\label{Lq1}
	\begin{split}
	\mathcal{L}_{q} &= \sum\limits_{i=1}^{L} \frac{1}{\bar{N}_{i}}  \sum\limits_{j=1}^{N_{i}} M_{ij} \left( W_{ij} - Q_{i}(W_{ij}) \right)^{2}.
	\end{split}
	\end{equation}
	Since $f_{i}(x)$ is a symmetric distribution function, 
	we could simplify the quantization error $\mathcal{L}_{q}$ in \Cref{Lq1} as:
	\begin{equation}\label{Lq}
	\resizebox{.42\textwidth}{!}{$
	\displaystyle \mathcal{L}_{q} = 2 \sum\limits_{i=1}^{L} \frac{1}{\bar{N}_{i}} \int_{\beta_{i}}^{+\infty} N_{i} f_{i}(x) (x - Q_{i}(x))^{2} dx \approx \sum\limits_{i=1}^{L} \frac{\Delta_{i}^{2}}{12},
	$}
	\end{equation}
	where $\bar{N}_{i} = 2 N_{i} \int_{\beta_{i}}^{+\infty} f_{i}(x) dx$ is the number of unpruned weights in the $i$-th layer. 
	
	To minimize $\mathcal{L}_{q}$ with bit constraint, we introduce a multiplier $\lambda$~(\textit{i.e.}, a Lagrange multiplier) to enforce the bit-allocation requirement in \Cref{bits} as follows:
	\begin{equation}
	\begin{split}
	&\mathcal{L}(\Delta_{1}, \cdots, \Delta_{L}, \lambda) \\
	&~=\sum\limits_{i=1}^{L} \frac{1}{12} \Delta_{i}^{2} + \lambda \left( \frac{1}{\bar{N}} \sum_{i=1}^{L} \sum_{j = 1}^{C_{i}} \bar{N}_{ij} \frac{2\alpha_{ij}}{\Delta_{i}} - 2^{B} \right),
	\end{split}
	\end{equation}
	where $\bar{N}_{ij}$ is the number of unpruned connections in the $j$-th channel of the $i$-th layer. Similar to the Lagrange derivation for pruning in \Cref{pruning}, we apply Lagrange derivation to derive $\Delta_{i}$ and $\lambda$ as:
	\begin{equation}\label{Delta_i}
	\Delta_{i} = \sqrt[3]{\frac{12 \lambda \sum_{j = 1}^{C_{i}} \bar{N}_{ij} \alpha_{ij}}{\bar{N}}},
	\end{equation}
	\begin{equation}\label{lambda3}
	\lambda = \left( \frac{1}{2^{B-1} \bar{N}} \sum_{i = 1}^{L} \sqrt[3]{\frac{\bar{N}}{12}} \frac{\sum_{j = 1}^{C_{i}} \bar{N}_{ij} \alpha_{ij}}{\sqrt[3]{\sum_{j = 1}^{C_{i}}  \bar{N}_{ij}\alpha_{ij}} } \right)^{3}.
	\end{equation}
	Combining \Cref{Delta_i,lambda3}, we obtain the quantization quotas (\textit{i.e.}, $\{ \Delta_{i} \}_{i = 1}^{L}$) for all layers, which can be used to quantize the given DNN model in the forward pass.

	\section{Experiments}
	In this section, we evaluate our method by compressing well-known convolutional neural networks, \textit{i.e.}, AlexNet~\cite{krizhevsky2012imagenet}, VGG-16~\cite{simonyan2014very}, ResNet-50~\cite{he2016deep}, MobileNet-V1~\cite{howard2017mobilenets}. 
	All experiments are performed on ImageNet (\textit{i.e.}, ILSVRC-2012)~\cite{deng2009imagenet}, a large-scale image classification dataset consisted of 1.2M training images and 50K validation images. To demonstrate the advantage of our method, we compare our results with state-of-the-art pruning, quantization and in-parallel pruning-quantization methods. Finally, we conduct error analysis to verify the validity of our method.
	
	\subsection{Implementation Details}
	The proposed method is implemented by PyTorch. We set the batch size as 256 for all models at finetuning stage. The SGD otpimizer is utilized to finetune the compressed models with the momentum ($=0.9$), weight decay ($=10^{-4}$), and learning rate ($=0.005$). Since pruning has a greater impact on performance than quantization, we adopt two-stage strategy to finetune the compressed networks. First, we finetune the pruned networks without quantization till they recover the performance of the original uncompressed models. Second, pruning and quantization are simultaneously applied at the finetuning stage.
	
	\begin{table*}[!thb]
		\centering
		\resizebox{.98\textwidth}{!}{ %
			\begin{tabular}{llllll}
				\hline\hline
				Method                                   & Top-1 (\%)                    & Top-5 (\%)                    & Prune Rate (\%) & Bit  & Rate           \\ \hline
				Data-Free Pruning~\cite{srinivas2015data}       & 55.60 (2.24$\downarrow$) & -                        & 36.24 & 32   & 1.57$\times$   \\
				Adaptive Fastfood 32~\cite{yang2015deep}         & \textbf{58.10} (0.69$\uparrow$)   & -                        & 44.12 & 32   & 1.79$\times$   \\
				Less Is More~\cite{zhou2016less}                 & 53.86 (0.57$\downarrow$) & -                        & 76.76 & 32   & 4.30$\times$   \\
				Dynamic Network Surgery~\cite{guo2016dynamic}    & 56.91 (0.3$\uparrow$)    & 80.01 (-)                   & 94.3  & 32   & 17.54$\times$  \\
				Circulant CNN~\cite{cheng2015exploration}        & 56.8 (0.4$\downarrow$)   & \textbf{82.2} (0.7$\downarrow$)   & 95.45  & 32   & 18.38$\times$  \\
				Constraint-Aware~\cite{chen2018constraint}       & 54.84 (2.57$\downarrow$)   & -                        & \textbf{95.13} & 32   & 20.53$\times$  \\
				Q-CNN~\cite{wu2016quantized}                     & 56.31 (0.99$\downarrow$)  & 79.70 (0.60$\downarrow$) & -     & 1.57 & 20.26$\times$  \\
				Binary-Weight-Networks~\cite{rastegari2016xnor}  & 56.8 (0.2$\uparrow$)   & 79.4 (0.8$\downarrow$)   & -     & \textbf{1}    & 32$\times$     \\
				Deep Compression~\cite{han2015deep}              & 57.22 (0.00$\uparrow$)   & 80.30 (0.03$\uparrow$)   & 89    & 5.4  & 6.66$\times$   \\
				CLIP-Q~\cite{tung2020deep}                                   & 57.9 (0.7$\uparrow$)     & -                        & 91.96 & 3.34 & 119.09$\times$ \\
				ANNC~\cite{yang2020automatic}                                     & 57.52 (\textbf{1.00$\uparrow$})   & 80.22 (0.03$\uparrow$)   & 92.6  & 3.7  & 118$\times$    \\ \hline
				Ours                                     & 57.09 (0.46$\uparrow$)   & 80.25 (\textbf{1.20$\uparrow$})   & 92.30  & 2.99  & \textbf{138.96$\times$}     \\\hline\hline
			\end{tabular}
		}
		\caption{AlexNet on ImageNet.}\label{table:alexnet}
	\end{table*}
	
	\begin{table*}[!thb]
		\centering
		\begin{tabular}{llllll}
			\hline\hline
			Method                                                                      & Top-1 (\%)                          & Top-5 (\%)                          & Prune Rate (\%)       & Bit           & Rate \\
			ThiNet-GAP~\cite{luo2017thinet}                           & 67.34 (1.0$\downarrow$)        & 87.92 (0.52$\downarrow$)        & 94.00          & 32             & 16.63$\times$   \\
			Q-CNN~\cite{wu2016quantized}                   & 68.11 (3.04$\downarrow$)          & 88.89 (1.06$\downarrow$)          & -              & \textbf{1.35}            & 23.68$\times$      \\
			Deep Compression~\cite{han2015deep}                   & 68.83 (0.33$\uparrow$)        & 89.09 (\textbf{0.41}$\uparrow$)          & 92.5              & 6.4             & 66.67$\times$       \\
			CLIP-Q~\cite{tung2020deep}                                                                      & 69.2 (\textbf{0.7$\uparrow$})            &   -                           & 94.20          & 3.06 & 	180.47$\times$  \\\hline			
			Ours                                                                        & \textbf{71.39} (0.24$\downarrow$) & \textbf{90.28} (0.09$\downarrow$) & \textbf{94.41} & 2.92          & \textbf{195.87$\times$} \\\hline\hline
		\end{tabular}
		\caption{VGG-16 on ImageNet.}\label{table:vgg}
	\end{table*}
	
	\subsection{Comparisons with State-of-the-Art}
	We performed extensive experiments to evaluate the effectiveness of the proposed method by comparing with 12 state-of-the-art methods, including 6 pruning methods~(\textit{i.e.}, \textit{i.e.}, Data-Free Pruning~\cite{srinivas2015data}, Adaptive Fastfood 32~\cite{yang2015deep}, Less Is More~\cite{zhou2016less}, Dynamic Network Surgery~\cite{guo2016dynamic}, Circulant CNN~\cite{cheng2015exploration}, and Constraint-Aware~\cite{chen2018constraint}), 3 quantization methods~(\textit{i.e.}, Q-CNN~\cite{wu2016quantized}, Binary-Weight-Networks~\cite{rastegari2016xnor}, and ReNorm~\cite{he2018learning}), 3 pruning-quantization methods~(\textit{i.e.}, Deep Compression~\cite{han2015deep}, CLIP-Q~\cite{tung2020deep}, and ANNC~\cite{yang2020automatic}).
	
	\subsubsection{AlexNet on ImageNet}
	For AlexNet, we compare our method with pruning only~(\textit{i.e.}, Data-Free Pruning~\cite{srinivas2015data}, Adaptive Fastfood 32~\cite{yang2015deep}, Less Is More~\cite{zhou2016less}, Dynamic Network Surgery~\cite{guo2016dynamic}, Circulant CNN~\cite{cheng2015exploration}, and Constraint-Aware~\cite{chen2018constraint}), quantization only~(\textit{i.e.}, Q-CNN~\cite{wu2016quantized} and Binary-Weight-Networks~\cite{rastegari2016xnor}) and pruning-quantization~(\textit{i.e.}, Deep Compression~\cite{han2015deep}, CLIP-Q~\cite{tung2020deep}, and ANNC~\cite{yang2020automatic}) methods. Unlike our one-shot pruning-quantization framework, all the compared methods gradually learn their pruning masks and/or quantizers during fine-tuning process. 
	Our method compresses the pre-trained uncompressed AlexNet which achieves 56.62\% top-1 accuracy and 79.06\% top-5 accuracy. Experimental results are shown in \Cref{table:alexnet}. From the table, we observe that our method compresses AlexNet to be 138.96$\times$ smaller while preserving the accuracy of the uncompressed AlexNet on ImageNet. Our method could even achieve 0.46\% top-1 and 1.20\% top-5 accuracy improvements. Compared to ANNC with 118x compression rate, our method performs slightly lower in Top-1 and slightly higher in Top-5, but with over 3x less finetuning time (160h vs 528h).
	
	\subsubsection{VGG-16 on ImageNet}
	For VGG-16, we compare our method with pruning only~(\textit{i.e.}, ThiNet-GAP~\cite{luo2017thinet}), quantization only~(\textit{i.e.}, Q-CNN~\cite{wu2016quantized}) and pruning-quantization~(\textit{i.e.}, Deep Compression~\cite{han2015deep}, and CLIP-Q~\cite{tung2020deep}) methods. Our method compresses the pre-trained uncompressed VGG-16 of PyTorch which achieves 71.63\% top-1 and 90.37\% top-5 accuracy. Experimental results are shown in \Cref{table:vgg}, one can see that our method can compress VGG-16 to be 195.87$\times$ smaller with only negligible accuracy loss comparing with the uncompressed model. Compared to other approaches, out method still achieves the best accuracy with the highest compression rate.
	
	

	\subsubsection{MobileNet-V1 on ImageNet}
	For MobileNet-V1, we compare our method with pruning only~(\textit{i.e.}, To Prune or Not To Prune~\cite{zhu2017prune}), quantization only~(\textit{i.e.}, Deep Compression~\cite{han2015deep}, ReNorm~\cite{he2018learning}, and HAQ~\cite{wang2019haq}) and pruning-quantization~(\textit{i.e.}, CLIP-Q~\cite{tung2020deep} and ANNC~\cite{yang2020automatic}) methods. 
	A pre-trained uncompressed MobileNet-V1 model, which achieves 70.28\% top-1 and 89.43\% top-5, is adopted for pruning and quantization. \Cref{table:mobilenet} shows the experimental results comparing with the aforementioned compression methods. From the table, we can see that our method is superior to all the other methods with better accuracy and higher compression rate. Specifically, our method achieves a 0.55\% top-1 accuracy improvement~(\textit{i.e.}, 70.83\%) with compressing rate $23.26\times$, while CLIP-Q~\cite{tung2020deep} with no improvement in top-1 accuracy at a much lower compression rate $13.19\times$. Our method also outperforms the most recent in-parallel pruning-quantization method ANNC~\cite{yang2020automatic} by a large margin in terms of accuracy improvement and compression rate.
	
	\begin{table*}[!thb]
		\centering
		\begin{tabular}{llllll}
			\hline\hline
			Method                        & Top-1 (\%)                         & Top-5 (\%)                         & Prune Rate (\%)          & Bit           & Rate             \\ \hline
			To Prune or Not To Prune~\cite{zhu2017prune}      & 69.5 (1.1$\downarrow$)         & 89.5 (0.0$\uparrow$)                              & 50           & 32            & 2$\times$                       \\
			Deep Compression~\cite{han2015deep}                   & 65.93 (4.97$\downarrow$)          & 86.85 (3.05$\downarrow$)          & -              & 3             & 10.67$\times$      \\
			ReNorm~\cite{he2018learning}                   & 65.93 (9.75$\downarrow$)          & 83.48 (6.37$\downarrow$)          & -              & 4             & 8$\times$      \\
			HAQ~\cite{wang2019haq}                                & 67.66 (3.24$\downarrow$)        & 88.21 (1.69$\downarrow$)          & -              & 3             & 10.67$\times$       \\
			CLIP-Q~\cite{tung2020deep}                        & 70.3 (0.0$\uparrow$)           & -                           & 47.36          & 4.61          & 13.19$\times$          \\
			ANNC~\cite{yang2020automatic}                          & 69.71 (1.19$\downarrow$)        & 89.14 (0.76$\downarrow$)          & -          & 3          & 10.67$\times$          \\ 
			ANNC~\cite{yang2020automatic}                          & 66.49 (4.41$\downarrow$)        & 87.29 (2.61$\downarrow$)          & 58          & \textbf{2.8}          & 26.7$\times$          \\ \hline
			Ours                          & \textbf{70.83 (0.55$\uparrow$)} & \textbf{89.70 (0.27$\uparrow$)} & 57.78 & 3.26 & 23.26$\times$ \\
			Ours                          & 70.24 (0.04$\downarrow$) & 89.30 (0.13$\downarrow$) & \textbf{67.66} & 3.08 & \textbf{32.15$\times$} \\\hline\hline
		\end{tabular}
		\caption{MobileNet-V1 on ImageNet.}\label{table:mobilenet}
	\end{table*}

	\subsubsection{ResNet-50 on ImageNet}
	\begin{table*}[!thb]
		\centering
		\begin{tabular}{llllll}
			\hline\hline
			Method                                                                      & Top-1 (\%)                           & Top-5 (\%)                           & Prune Rate (\%)          & Bit           & Rate \\
			ThiNet~\cite{luo2017thinet}                           & 71.01 (1.87$\downarrow$)        & 90.02 (1.12$\downarrow$)        & 51.76          & 32             & 2.07$\times$   \\
			Deep Compression~\cite{han2015deep}                   & 76.15 (0.00$\uparrow$)          & 92.88 (0.02$\uparrow$)          & -              & 4             & 8$\times$      \\
			HAQ~\cite{wang2019haq}                                & 76.14 (0.01$\downarrow$)        & 92.89 (0.03$\uparrow$)          & -              & 4             & 8$\times$       \\
			ACIQ~\cite{banner2019post}                              & 75.3 (0.8$\downarrow$)        & -          & -              & 4             & 8$\times$       \\
			CLIP-Q~\cite{tung2020deep}                                                                      & 73.7 (\textbf{0.6$\uparrow$})            &   -                           & 69.38          & 3.28 & 31.81$\times$  \\\hline
			Ours                                                                        & \textbf{76.41} (0.40$\uparrow$) & \textbf{93.04 (0.11$\uparrow$)} & \textbf{74.14} & \textbf{3.25}          & \textbf{38.03$\times$} \\\hline\hline
		\end{tabular}
		\caption{ResNet-50 on ImageNet.}\label{table:resnet}
	\end{table*}
	For ResNet-50, we compare our method with pruning only~(\textit{i.e.}, ThiNet~\cite{luo2017thinet}), quantization only~(\textit{i.e.}, Deep Compression~\cite{han2015deep}, HAQ~\cite{wang2019haq}, and ACIQ~\cite{banner2019post}) and pruning-quantization~(\textit{i.e.}, CLIP-Q~\cite{tung2020deep}) methods. 
	Our method compresses a pre-trained uncompressed ResNet-50 model which reports 76.01\% top-1 and 92.93\% top-5 on ImageNet. \Cref{table:resnet} shows the experimental results of our method comparing with the aforementioned state-of-the-art methods. From the table, we can see that our method could compress ResNet-50 to be 38.03$\times$ smaller with 0.40\% top-1 and 0.11\% top-5 improvements. Comparing to all the other methods, out method achieves the best accuracy~(76.41\% top-1 and 93.04\% top-5) with the highest compression rate, while without gradually learning the pruning masks and quantizers at finetuning stage.
	
	\subsubsection{Summary}
	In conclusion, extensive experimental results on ImageNet with various neural network architectures show that in-parallel pruning-quantization boosts the model compression rate by a large margin without incurring accuracy loss, compared to pruning only and quantization only approaches. More importantly, our One-shot Pruning-Qunatization consistently outperforms the other in-parallel pruning-quantization approaches which gradually optimize the pruning masks and quantizers at finetuning stage, suggesting that pre-trained model is sufficient for determining the compression module prior to finetuning, and the complex and iterative optimization of the compression module may not be necessary at finetuning stage.

	\subsection{Error Analysis}
\begin{figure}[!thb]
		\begin{subfigure}[b]{0.46\linewidth}
			\includegraphics[height=0.95\linewidth]{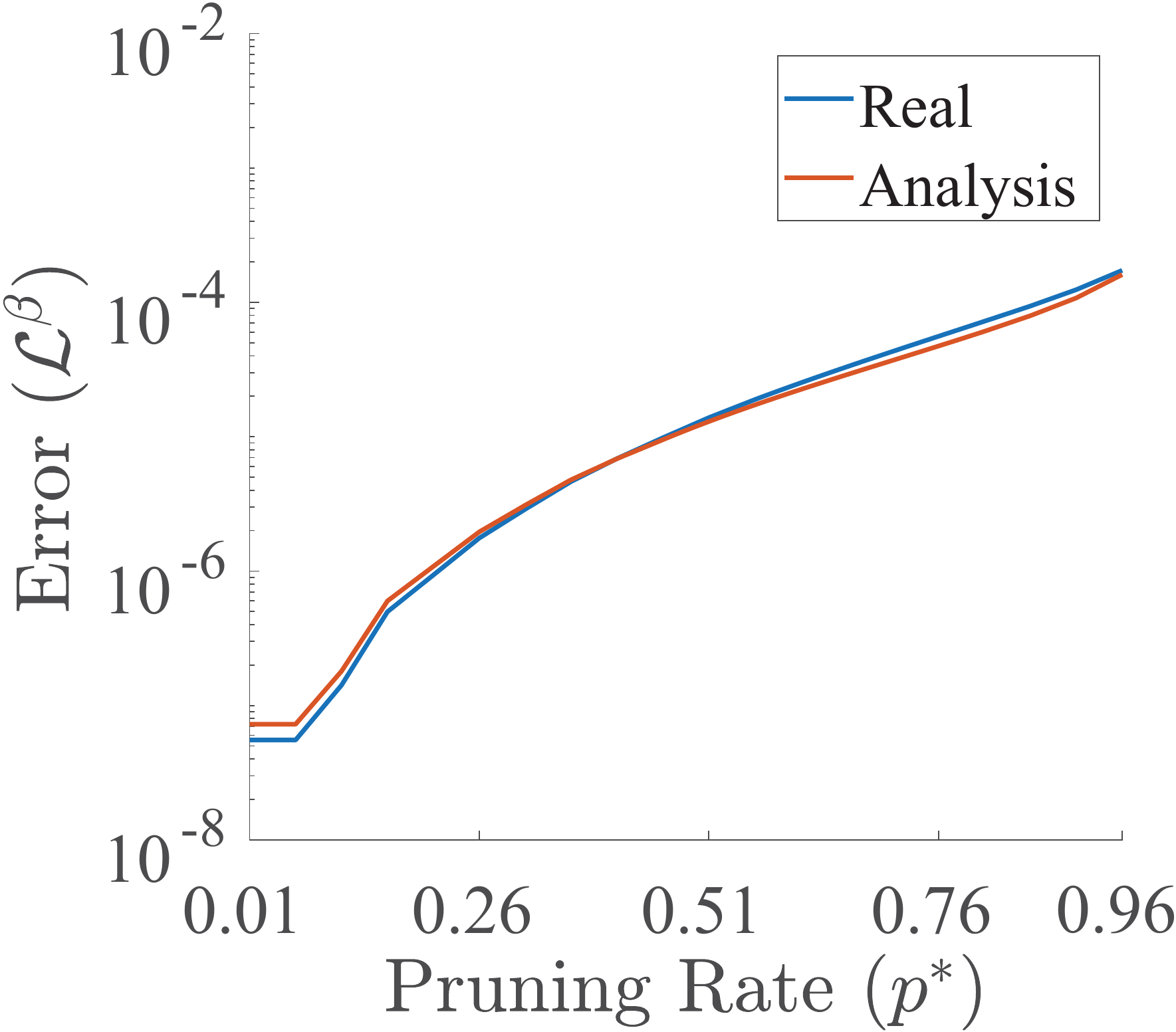}
			\caption{Pruning Error Analysis.}\label{fig:pruning_error}
		\end{subfigure}
		\hspace{2mm}
		\begin{subfigure}[b]{0.46\linewidth}
			\centering
			\includegraphics[height=0.95\linewidth]{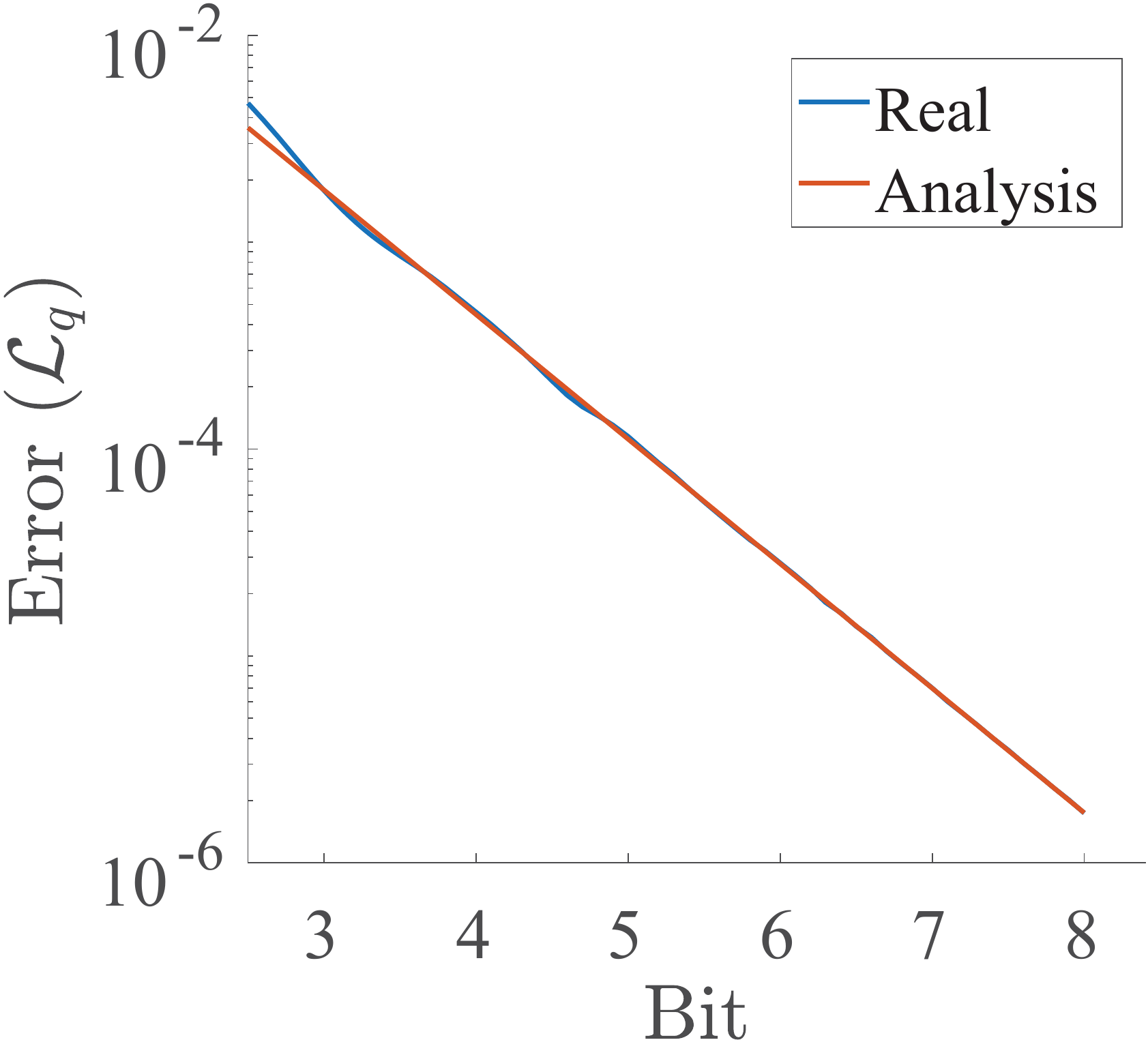}
			\caption{Quantization Error Analysis.}\label{fig:quantization_error}
		\end{subfigure}
		\caption{Error analysis on ResNet-50. (\subref{fig:pruning_error}) illustrates the difference between the analyzed and the real pruning errors. ``Real'' denotes the real pruning error, \textit{i.e.}, $\frac{1}{N} \sum_{i = 1}^{L} \sum_{j=1}^{N_{i}} \left( W_{ij} \right)^{2} \big\rvert_{|W_{ij}| \leqslant \beta_{i}}$ in \Cref{eq:obj}. ``Analysis'' denotes the approximated pruning error using Laplace probability density functions, \textit{i.e.}, $\frac{2}{N} \sum_{i=1}^{L} \int_{0}^{\beta_{i}} N_{i} x^{2} f_{i}(x)dx$, where $f_{i}(x) = \frac{1}{2\tau_{i}} e^{-\frac{|x|}{\tau_{i}}}$. (\subref{fig:quantization_error}) shows the difference between the analyzed and the real quantization errors. ``Real'' presents the real quantization error, \textit{i.e.}, \Cref{Lq1}. ``Analysis'' denotes the approximated quantization error, \textit{i.e.}, $\sum_{i=1}^{L} \frac{\Delta_{i}^{2}}{12}$ in \Cref{Lq}.}\label{fig:error_analysis}
	\end{figure}

	In this section, we investigate the relationships between the analytic errors and the real ones computed on ResNet-50. From \Cref{fig:pruning_error}, we can see that the Laplace probability density function approximates the real pruning error well, demonstrating the validity of \Cref{lambda2}. From \Cref{fig:quantization_error}, we observe that the approximation error $\sum_{i=1}^{L} \frac{\Delta_{i}^{2}}{12}$ is in a good agreement with the real quantization error of \Cref{Lq1}, which verifies the effectiveness of the proposed approximation. Therefore, the proposed method is able to capture the prior knowledge of the pre-trained models effectively, which in turn guides the design of our analytical solution for one-shot pruning-quantization.
	
	\section{Conclusion}
	In this paper, we propose a novel One-shot Pruning-Quantization method~(OPQ) to compress DNN models. Our method has addressed two challenging problems in network compression. First, different from the prior art, OPQ is a one-shot compression method without manual tuning or iterative optimization of the compression strategy~(\textit{i.e.}, pruning mask and quantization codebook of each layer). The proposed method analytically computes the compression allocation of each layer with pre-trained weight parameters only. Second, we propose a unified channel-wise pruning to enforce all channels of each layer to share a common codebook, which avoids the overheads brought by the traditional channel-wise quantization. Experiments show that our method achieves superior results comparing to the state-of-the-art. For AlexNet, MobileNet-V1, ResNet-50, our method could improve accuracy even at very high compression rates. For future work, we will explore how to further compress DNN models, and implement our method with custom hardware architecture in order to validate the inference efficiency of the compressed models on practical hardware platforms.
	
	\section*{Acknowledgments}
	This work is supported in part by the Agency for Science, Technology and Research (A*STAR) under its AME Programmatic Funds (Project No.A1892b0026, A18A2b0046 and Project No.A19E3b0099); the Fundamental Research Funds for the Central Universities under Grant YJ201949; and the NFSC under Grant 61971296, U19A2078, and 61836011.
	
	\section*{Broader Impact Statement}
	Neural network compression is a hot topic which has drawn tremendous attention from both academia and industry. The main objective is to reduce memory footprint, computation cost and power consumption of the training and/or inference of deep neural networks, and thus facilitate their deployments on resource-constrained hardware platforms, \textit{e.g.}, smartphones, for a wide range of applications including computer vision, speech and audio, natural language processing, recommender systems, etc.
	
	In the paper, we mainly focus on exploring the one-shot model compression mechanism, in particular in-parallel pruning-quantization, for model compression at extremely high compression rate without incurring loss in accuracy. Our One-shot Pruning-Quantization method (OPQ) largely reduces the complexity for optimizing the compression module and also provides the community new insight into how can we perform efficient and effective model compression from alternative perspective.
	Although OPQ achieves promising performance in network compression, we should also care about the potential negative impacts including 1) the compression bias caused by OPQ because of unusual weight distribution, too lower objective compression rate, etc. Usually, this requires domain experts to manually evaluate them. 2) the robustness of the compressed models for decision making, especially in health care, autonomous vehicles, aviation, fintech, etc. Question to be answered could be is the compressed model vulnerable to adversarial attacks because of the introduction of prunining and quantization into the model? 
	We encourage further study to understand and mitigate the biases and risks potentially brought by network compression. 
	
	\bibliography{Bibliography}
\end{document}